\documentclass[fleqn,10pt]{wlpeerj}
\usepackage{tcolorbox}
\usepackage{float}
\usepackage[ruled, lined, linesnumbered,  longend]{algorithm2e}
\SetKwInput{KwInput}{Input}                
\SetKwInput{KwOutput}{Output}              
\title{AdCOFE: Advanced Contextual Feature Extraction in Conversations for emotion classification}

\author{Vaibhav Bhat}
\author{Anita Yadav}
\author{Sonal Yadav}
\author{Dhivya Chandrasekran}
\author{Vijay Mago}
\affil{Department of Computer Science, Lakehead University, Ontario, Canada}

\begin{abstract}
Emotion recognition in conversations is an important step in various virtual chat bots which require opinion-based feedback, like in social media threads, online support and many more applications. Current Emotion recognition in conversations models face issues like (a) loss of contextual information in between two dialogues of a conversation, (b) failure to give appropriate importance to significant tokens in each utterance and (c) inability to pass on the emotional information from previous utterances.The proposed model of Advanced Contextual Feature Extraction (AdCOFE) addresses these issues by performing unique feature extraction using knowledge graphs, sentiment lexicons and phrases of natural language at all levels (word and position embedding) of the utterances. Experiments on the Emotion recognition in conversations dataset show that AdCOFE is beneficial in capturing emotions in conversations.
\end{abstract}

\begin{document}

\flushbottom
\maketitle
\thispagestyle{empty}

\section*{Introduction}

There has been a significant advancement in conversational AI research, where the emotion recognition in conversations has been acknowledged as one of the crucial areas of research. The task of identifying or predicting the emotion of each dialogue in any conversation is called Emotion recognition in conversations (ERC) \citep{Hazarika2021}. This identification or prediction of emotions in conversation is an essential step in any conversation understanding applications. ERC has proven to be indisputably important in real time dialogue systems like emotion-aware chat agents \citep{Li2020}, visual question answering \citep{Tapaswi2016} and many other applications where the potential of ERC is immense.

Various virtual digital agents and chat bots are available on different websites which are used by a large number of online users. These chat bots rely mainly on the ability to generate responses depending upon the user emotions. The bots need to detect these emotions and sentiments in order to provide emotionally coherent and empathetic responses by avoiding any inappropriate response \citep{Miner2016}. In conversational agents, models are trained to understand human emotions in conversational context \citep{Zhong2020}. Emotion detection has also been providing effective results in the area of opinion mining over a chat history, social media threads and debates. 

There is a wide range of other applications which can utilize ERC. For instance, while providing customer service on a social media platform like Twitter, ERC models can provide quick responses to the issues raised through numerous tweets \citep{Chatterjee2019}. Models can prioritize the tweets according to the emotion in the content and respond to the messages with emotion class of angry or upset providing them maximum user satisfaction \citep{Chatterjee2019}. Additionally, ERC can help applications to prevent sending or posting of messages by tagging them as bullying or threatening, thus warning online users before sending such messages.

The different ERC models proposed in the existing research provide the sentiment classification of the utterances in a dyadic conversation (a conversation of two participants). These models have not captured some of the key parameters which needs to be attended in order to recognize emotion in conversations. Some models do not perform ideal feature extraction, or some models lack the minute attention which must be given to significant tokens in an utterance. In AdCOFE, we intend to address such issues by extracting advanced contextual features in the feature extraction layer prior to passing it through the main classification model. Key methodological highlights of our research work are: 
(a) Utilizing knowledge graphs for advanced contextual understanding of the sentences, (b) Employment of sentiment lexicons for adding context-based emotional features to the sentences, and (c) Use of simple and elegant pretrained model which is efficiently fine-tuned for the purpose of ERC. With these advances, the results achieved are better as compared to the state-of-the-art models, and are comparable to the  latest models on the Interactive Emotional Dyadic Motion Capture (IEMOCAP) dataset.

The remainder of the paper is organized as follows: Related Works section provides insights of previous research work and the progress in ERC; Methodology section describes the dataset used, discusses our proposed approach, algorithm and provides details on experimental setup; Results and Discussion section reports the results and provides a comparative analysis with both the baseline as well as the latest ERC models; finally, we conclude the paper signifying the distinctive approach implemented and also mention the future work to further enhance the classification model.

\section*{Related Work}
Emotion recognition in conversation has gained immense popularity in a very small span of time due to its widespread applications. As described in the introduction section, ERC mainly consists of context representation, capturing the impact of the contextual information from previous utterances and extracting emotional features to perform classification. To achieve this contextual modeling in either textual or multimodal setting different deep learning-based algorithms are being used. \cite{Poria2017} used RNNs for multimodal emotion recognition which propagated contextual and sequential information to the utterances. \cite{Majumder2018} enhanced these deep learning models further by incorporating party and global states in recurrent model for modeling emotional dynamics. The party states are used to keep track of each speaker while the global state keeps account of the context. \cite{Ghosal2020} utilized the speaker-level context encoding like \cite{Majumder2018}, using a two-step graph convolution process. Along with speaker context, sequential context was extracted to overcome the issue of context propagation in the DialogueRNN model \citep{Majumder2018}. However, RNNs fail to take into consideration the dependencies between two utterances in a conversation causing loss of long-range contextual information in a dialogue. 

\cite{Jiao2019} proposed a hierarchical Gated Recurrent Unit (GRU) framework with self-attention and feature fusion (HiGRU-sf) model to capture long-range contextual information among words and utterances for ERC. In the further research \cite{Jiao2019a} proposed an Attention Gated Hierarchical Memory Network (AGHMN) where the Hierarchical Memory Network (HMN) is responsible for enhancing features at utterance level and memory bank that can be used for contextual information extraction. The two-level GRU layer of HMN is responsible for modeling the word sequence of each utterance called the utterance reader, and the other layer adopts a BiGRU structure to capture historical utterances. In addition to this HMN network, an Attention GRU (AGRU) is added which promotes comprehensive context and retains positional and ordering information of the utterance. \cite{Li2020} created a generalized neural tensor block (GNTB) followed by two emotion feature extractor (EFE) channel classifiers to perform context compositionality and sentiment classification respectively. Similar to bidirectional long short term memory (BiLSTM), two emotional recurrent units (ERU) are utilized for forward and backward passing of the input utterances after which the outputs from these forward and backwards ERUs are concatenated for sentiment classification or regression \citep{Li2020}. 

Additionally, the existing models only capture the contextual information of utterances in a single conversation and do not consider that each token in an utterance has different importance to represent the meaning of the utterance. This means that these models gave equal importance to each token in an utterance. \cite{Ragheb2019} suggested an attention-based modeling process to address the problem of equal weight to each token in a sentence. \cite{Ragheb2019} utilized three layered BiLSTM model, which was trained using Average Stochastic Gradient Descent (ASGD), and a self attention layer to extract features enriched with contextual information. Lately, \cite{9128015} proposed an Adapted Dynamic Memory Network (A-DMN) which models self and inter-speaker influences individually. The model further synthesizes the influence towards current utterance which was not incorporated effectively in previous models. Overall, due to the sequential nature of the data, the use of recurrent neural networks is in abundance. Moreover, it is pivotal that the speaker context, sequential context and the emotions of the previous utterances are passed throughout the model. 

Recently, there have been development in concept of utilizing Graphs to provide solution for recognizing emotions. \cite{Xu2020}, proposed EmoGraph which utilizes graph networks to identify the dependencies among various emotions. The co-occurrence statistics in between different emotion classes are used to create graphical neural network, taking each emotion class as a node. This helps the network in extracting features from the neighbours of each emotion node. \cite{Ishiwatari2020} highlighted that the different graph-based neural networks miss out on the sequential information of the conversation. To overcome this issue, a relational position encoding is proposed in the Relational Graph Attention Network (RGAT) which would reflect the speaker dependency and sequential information by applying positional encoding in the relational graph structure.

Another approach to capture emotions in conversation is by transfer learning model which utilizes a pre-trained hierarchical generative dialogue model, trained on multi-turn conversations \citep{hazarika-etal-2018-conversational}. The knowledge acquired from dialogue generators is transferred to a classification target model. The classification model uses the learned context and parameters from the pre-trained model output and projects it to the label-space \citep{hazarika-etal-2018-conversational} which helps it to identify the emotion for the given sentence. Furthermore, \cite{Zhong2020} used a Knowledge-Enriched Transformer for transfer learning using pre-trained weights. In the suggested model, all the tokens in the given utterance of the conversation is converted into a specific vector called Valence Arousal Dominance (VAD) \citep{Mohammad2018}. The conceptual knowledge is characterised using Dynamic Context-Aware Affective Graph Attention and the contextual knowledge is obtained using a Hierarchical self-attention. Moreover, a context-aware concept-enriched \citep{bahdanau2014neural} response representation for a given conversation is learned. Different pre-trained models and transformers have been implemented in the field of ERC to increase the computational speed as well as getting better results. Also, the usage of attention-based models in ERC is evident.

\section*{Methodology}
\subsection{Dataset}
The IEMOCAP dataset collected at SAIL lab at University of Southern California is designed to capture expressive human communication and synthesize different aspects of human behaviour \citep{Busso2008}. The dataset outlines a dyadic conversation among 10 different speakers. The dataset is an acted, multimodal database which contains 12 hours of audio-visual data including speech, motion capture of face and text transcriptions. The recorded sessions containing the dialogues were manually segmented based on the dialogues turn level and multi sentence utterances were split as single turns. The emotion labels assigned to the corpus were done using different annotation schemes which captured the emotional aspect of the dialogues.The most popular assessment schemes used were: discrete categorical based annotations (i.e., labels such as happiness, anger, and sadness), and continuous attribute-based annotations (i.e., activation, valence and dominance). Six human evaluators were asked to asses the emotional categories for the corpus. The interactions recorded were to be as natural as possible, hence the evaluators came across more than the normal set of emotions which led to ambiguous results. The final trade-off to classify the emotions of each utterance were categorized into the following emotions: anger, excitement, sadness, happiness, frustration and neutral states. Alternative approach to describe the emotional content of an utterance was to use primitive attributes such as valence, activation (or arousal), and dominance. But these categorical levels did not provide information about intensity level of the emotions. Therefore, having both types of emotional descriptions provided complementary insights about how people display emotions and how these cues can be automatically recognized or synthesized for better human-machine interfaces. This corpus is widely used in understanding expressive human communications and contributes to a better emotional recognition system.

\subsection{Problem Statement}
Let there be $n$ utterances in a conversation ($C$) represented as $u_n$. Let the two speakers in the dyadic conversation be denoted by \(p_1\) and \(p_2\). Let there be a mapping $S$ between the speaker and utterance denoted by $S:p_i \rightarrow u_j$ where $i \in \{0,1\}$ and $j \in [1,n]$ indicating which speaker uttered the given sentence in the conversation. We have a set of emotions $E$, which has values for $6$ emotions (Happy, Sad, Neutral, Angry, Excited and Frustrated) ranging from $0-5$. Our task is to identify the emotion of each mapping $S$ in the conversation.

\subsection{Data Preprocessing}
We use the IEMOCAP dataset for obtaining results from our proposed algorithm. The labels for each sentence are either of the following: Happy, Sad, Neutral, Angry, Excited, Frustrated for the training data \citep{Busso2008}. For preparing the data to pass through the respective layers of the model, two different processes of data preprocessing methods are used. At first, the data is preprocessed using the NLTK library to remove stop words and punctuation and then tokenized to extract contextual information using ConceptNet Web API \citep{Speer2016} and Python GET requests. This preprocessed data is further passed through the sentencepiece tokenizer and BERT \citep{Devlin2019} preprocessor to obtain a fine-tuned data for the {\it A Lite BERT} (ALBERT) \citep{Lan2019} pretrained model for classification.


\begin{figure*}[h!]
  \centering
  \begin{tabular}{@{}c@{\hspace{.1cm}}c@{}}
      \includegraphics[page=1,height =21.5cm, width=19.5cm,
      trim = -1cm 0cm -15cm 0cm,clip]{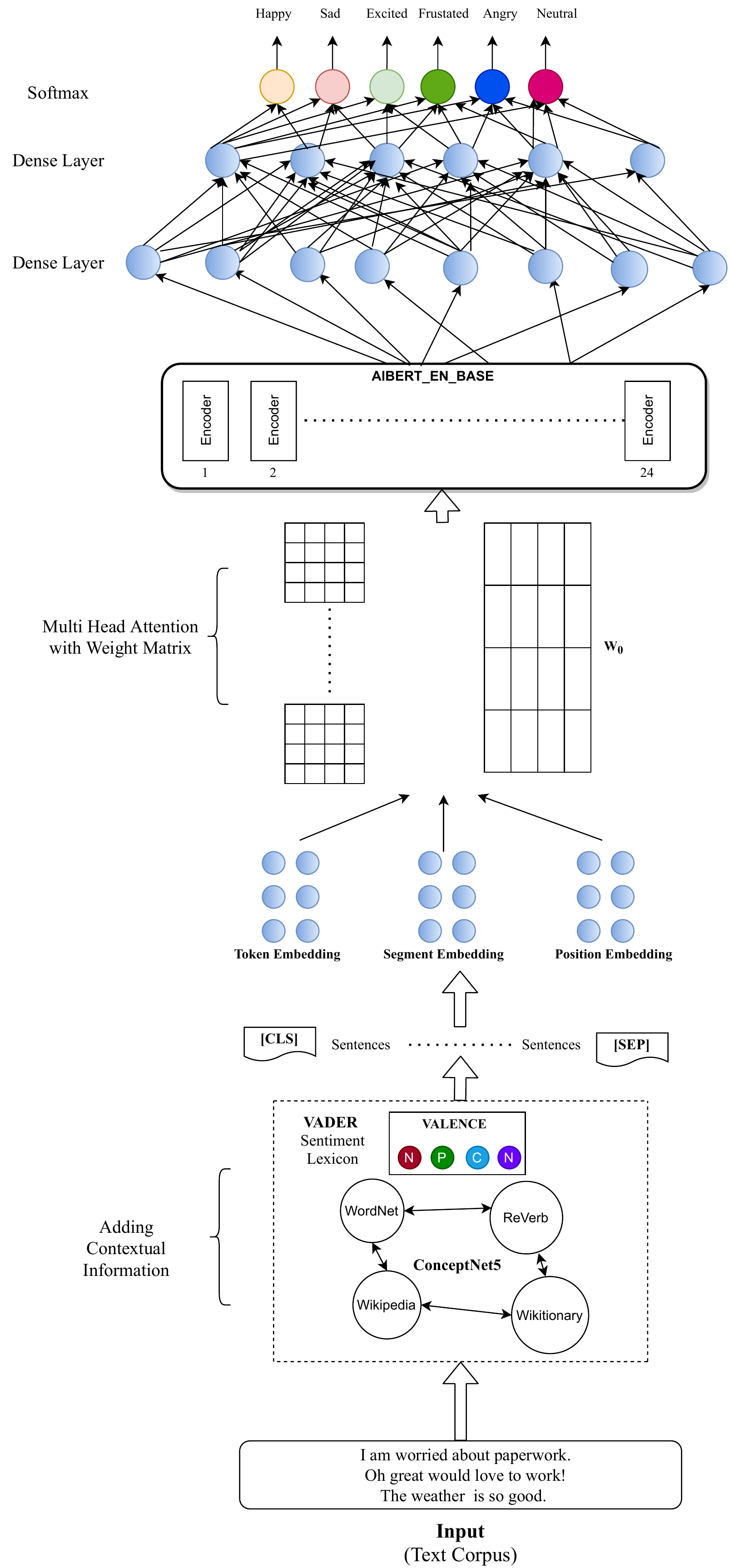}
  \end{tabular}
 \caption{Model Architecture}
 \label{fig:Model}
\end{figure*}

\subsection{Proposed Method}
Given transcripts of long range recorded dyadic conversations, ERC system aims to identify the emotions in each utterance and label them into appropriate emotion category. The major task here is to capture long range context from conversations and passing the emotions from one utterance to another. As the corpus consists of multiple words with different meanings based on the context, the task of the ERC system is to identify the emotion based on their position and importance in an utterance.

Keeping these shortcomings in mind, we propose AdCOFE\footnote{Implementation available at https://github.com/VBhat97/AdCOFE} which has the following characteristics:

\begin{itemize}
\item Context dependent feature extraction with equal importance to word  and position embedding in an utterance.
\item Contextual enrichment to existing sentences  by using knowledge graphs and emotion lexicons.
\item Incorporation of pretrained transformer models for final  emotion  classification.

\end{itemize}

The architecture in \figurename{~\ref{fig:Model}} summarizes the flow of our proposed method. The IEMOCAP corpus consists of long recorded dyadic conversations which serves as the input to our model.The initial step of AdCOFE is to maximize feature extraction for better classification. This is done by using basic preprocessing steps.In the next steps for contextual enhrichement we make use of ConceptNet API and VADER emotion lexicons which analyzes the context and adds on more to the existing corpus. These enriched sentences are then tokenized into various embeddings with special attentions to their positions in the utterance. The enhanced context enriched text corpus is then trained using the ALBERT model which then outputs a $1\times786$ dimensional matrix which is passed to the Fully Connected Dense Layers. At the end, we have a{\it softmax} layer which classifies the sentence according to one of the 6 emotion labels.

\subsubsection{Initial Feature extraction}
The text corpus of IEMOCAP dataset involves set of dyadic conversations which involves two participants. Each participant has a set of utterances which is associated with a particular emotion. The initial step of any processing involves data preprocessing or cleaning of data to remove unwanted words that could affect the emotion associated with the utterances which is done using the basic data cleaning techniques.After this the main focus is on context extraction which our model AdCOFE handles by passing the utterances from a single speaker in a batch of sentences. This step ensures that the emotion is carried throughout the model while keeping the length of conversations constant.This helps in easy classification of emotions even in long range conversations.


\begin{algorithm}[h]
\DontPrintSemicolon
\SetAlgoLined
\SetNoFillComment
\LinesNumbered
\KwInput{Dataset as $D_{train}$}
\KwOutput{Modified Dataset as $preprocessed\_D_{train}$}
$modified\_D_{train}$ $\gets$ arrange\_sentences($D_{train}$)\\ \tcp*{Arranging sentences to pass in a batch} 
$cleaned\_sentences$ $\gets$ remove\_punctuation($D_{train}$)\\ \tcp*{Remove Punctuation marks} 
$tokens \gets$ remove\_SW(Tokenize($cleaned\_sentences$))\\ \tcp*{Tokenize and Remove stop words}
$cpt\_sentences \gets$  ConceptNetAPI.get($tokens$) \\ \tcp*{Call ConceptNetAPI with the tokens}
$modified\_D_{train} \gets$ $ modified\_D_{train}+ cpt\_sentences$\\
$preprocessed\_D_{train}$ $\gets$ addVADERFeatures($modified\_D_{train}$)\\
\tcp*{Adding VADER Features}
\Return $preprocessed\_D_{train}$
\caption{Pseudocode for pre-processing the Data}
\label{algo: AlgorithmPre}
\end{algorithm}

\begin{figure*}[h!]
  \centering
  \begin{tabular}{@{}c@{\hspace{.1cm}}c@{}}
      \includegraphics[page=1,height =7.5cm, width=19cm,
      trim = -1cm 9cm -6cm 0.95cm,clip]{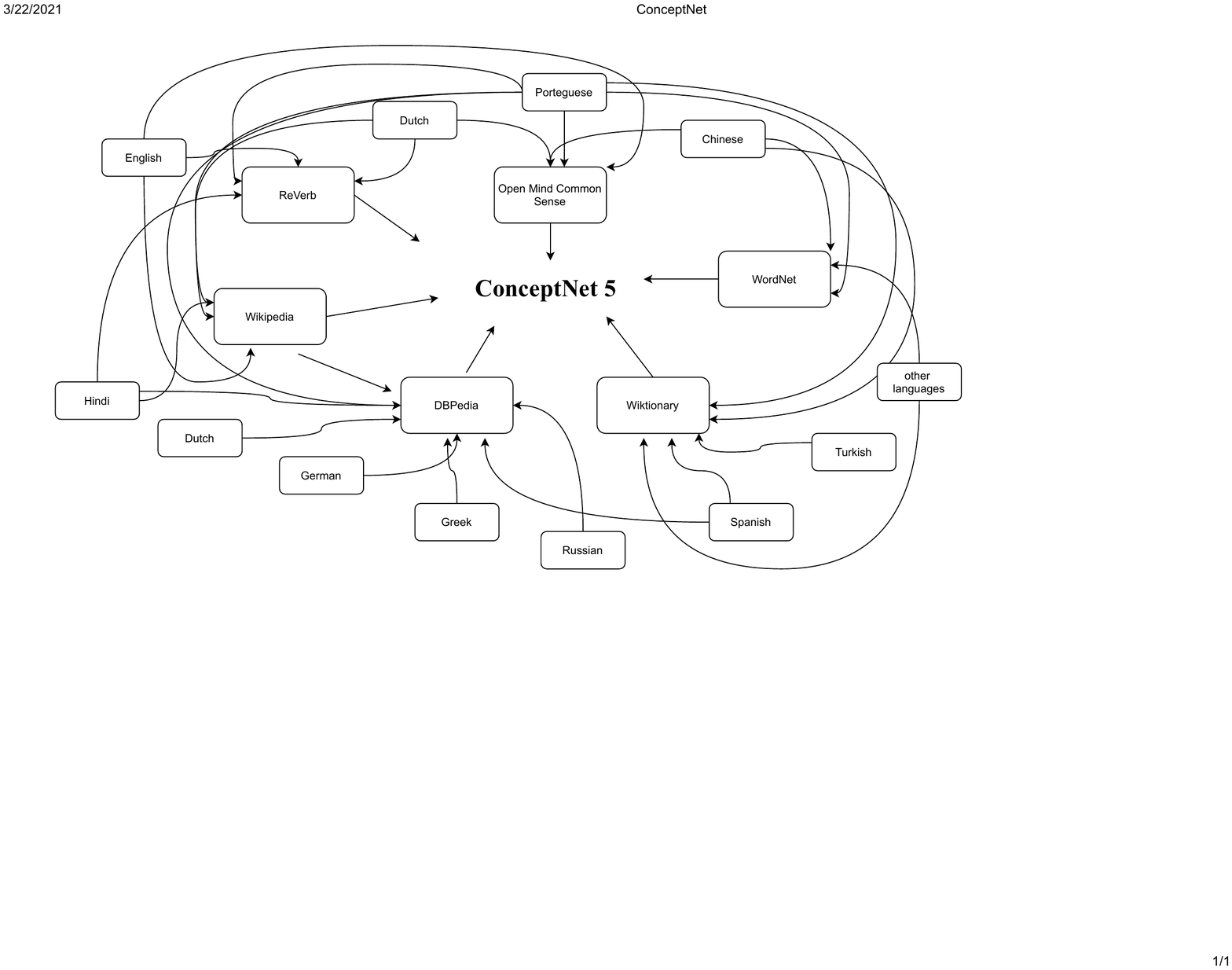}
  \end{tabular}
 \caption{ConceptNet Graph}
 \label{fig:Graph}
\end{figure*}

The next steps deals with context enrichment to already existing text corpus. We employ a ConceptNet model as proposed by \cite{Speer2016} (example shown in \figurename{~\ref{fig:Graph}} ) which  is a knowledge graph that connects words and phrases of natural language (terms), weighted edges (assertions) \cite{Speer2016}. For our model we use ConceptNet 5.7 to extract connected words to the given words in the utterances. ConceptNet Web API is used for getting the required contextual information through GET requests. After arranging the sentences as mentioned in the previous step, each sentence is preprocessed using the ConceptNet preprocessing as explained in Algorithm \ref{algo: AlgorithmPre}. The tokens are then passed to the ConceptNet Web API to retrieve the top related words as response to the request based on the ConceptNet knowledge graph. These tokens are then replaced in the sentences and added to the training data. An example is shown in the box below:

\medskip
\begin{tcolorbox}[colback=blue!5,colframe=blue!60!black,width=12cm, title=Example of ConceptNet contextual enrichment]
\textbf{Original Sentence}: This weather is the best!\\
\textbf{Contextual enrichment} : This \textless weathers\textgreater is the \textless bests\textgreater!\\
\textbf{Contextual enrichment} : The \textless climate conditions \textgreater is \textless besting \textgreater!\\
\textbf{Contextual enrichment} : The \textless weather \textgreater  wise is the \textless finest \textgreater!
\end{tcolorbox}
\medskip

The training data then contains both the original sentence as well as the contextually enriched sentences from ConceptNet. This enhances the contextual information in the training data by adding similar sentences.Thus, contextual features are obtained based on the knowledge graph of ConceptNet\footnote{ConceptNet API : http://api.conceptnet.io/}.

In the next step VADER \citep{Wilson2014}, a simple rule based model for sentiment analysis is used. The sentiment lexicon uses a combination of rules to generate lexical features which closely display the sentiment intensity of a particular sentence. The features obtained are rated on 4 scales from Negative, Positive, Neutral to Compound. These list of features are then used to determine the sentiment of the utterance based on a threshold score and are added with the context enriched sentences obtained in the previous step. The combination of contextual information obtained as a result of the ConceptNet API and the sentiment lexicon of VADER is then passed on to the ALBERT classification model.

\subsubsection{Emotion Classification Model}

For Emotion Classification, ALBERT is used for self-supervised learning of language representations. Results of ALBERT model are better as compared to the original BERT model {\citep{Devlin2019}} in ERC tasks. This is due to the appropriate architectural choices taken in the ALBERT model which helps to overcome the issues faced by the traditional ERC models. These specifically include:

\begin{itemize}
    \item \textbf{Factorized embedding parameterization:} The WordPiece embedding used in BERT \citep{Devlin2019}, XLNet \citep{Yang2019} and RoBERTa \citep{Liu2019} has a size that is tied with the size of the hidden layer. This is proved to be suboptimal from both modeling and practical perspective. From the modeling point of view, the untying of the size of WordPiece embeddings and hidden layer embeddings makes an efficient usage of total parameters, as in ERC these context-dependent representations have a great importance. From a practical point of view, due a large vocabulary of the sentences, the factorization of embedding parameters gives us an added advantage on the execution time as compared to BERT. Moreover, it also keeps in check the attention given to different words of a sentences which is pivotal as well.
    \item \textbf{Cross-layer parameter sharing:} ALBERT uses cross-layer parameter sharing strategy which helps on stabilizing the network parameters. In terms of ERC, this helps in finding the best options for tuning the hyperparameters much quicker as compared to BERT. Although it does not affect the hyperparameters such as number of epochs, learning rate, etc. this model architectural choice in ALBERT helps to reach them efficiently.
    \item \textbf{Inter-sentence coherence loss:} ALBERT follows sentence-order prediction (SOP) loss which does not take into consideration topic prediction  but is heavily based on inter-sentence coherence. This helps to improve the performance of multi-sentence encoding tasks. In terms of ERC, it becomes extremely important to take into account the coherence between the sentences as they are part of the same conversation. The three back-to-back utterances of a specific speaker is obviously connected and important for the speaker to be reasonable, thus taking into account coherence between the utterances plays an important role.  As a result, the combination of taking context from the words in an efficient way as well as the coherence in between the sentences together helps to select ALBERT over other BERT models.

\end{itemize}

\begin{algorithm}[h]
\DontPrintSemicolon
\SetAlgoLined
\SetNoFillComment
\LinesNumbered
\KwInput{Dataset as $preprocessed\_D_{train}$ and $train\_labels$, $D_{test}$ and $test\_labels$}
\KwOutput{Accuracy as \textit{A} and Weighted $F1$ Score as \textit{F1}}
$model$ $\gets$ pretrained\_model($``ALBERT\_en\_base"$)\\ \tcp*{Load ALBERT\_en\_base pretrained model} 
$no\_of\_layers \gets 3$ \\
\For{i in no\_of\_layers}{ 
    $model \gets $model + $ Dense(n=100; a\_f=``relu"; i)$ \\ \tcc{Adding 3 dense layers with 100 nodes and reLu activation function}
} 
$model \gets $model + $ Dense(n=6; a\_f=``softmax")$\\ \tcc{Adding Dense layer with 6 nodes and {\it softmax} activation function for output}
$epoch \gets 4$\\
\For{each e in epoch}{ 
    $utterance \gets preprocessed\_D_{train}(``batch\_size=1")$\\
    $h_{train}$ $\gets$ trainModel($utterance$)\\ \tcp*{Train classification Model}
}
$h_{test}$ $\gets$ runModel($D_{test}$, $h_{train}$)\\ \tcp*{Predicting for test data}
\textit{A} $\gets$ calculateAccuracy($test\_labels, h_{test}$) \\ 
$F1$ $\gets$ calculateF1Score($test\_labels, h_{test}$) \\ \tcp*{Calculating Accuracy and weighted F1 score}
\Return $A,F1$
\caption{Pseudo-code for Proposed Model}
\label{algo: Algorithm2}
\end{algorithm}

The proposed model, according to Algorithm \ref{algo: Algorithm2} uses a pre-trained ALBERT\_EN\_BASE model which serves the purpose of overcoming the shortcomings of an ERC model. The model has 12 attention heads, 12 hidden layers and 1 hidden group. The embedding size and hidden size are set to default with values 128 and 768 respectively. The size of maximum positional embeddings are 512. The vocabulary size is set to 30,000. The input to the model is similar to that given to a BERT model with input ids, masks and segment ids which are {\it int32} tensors of shape {\it batch\_size}, {\it max\_sequence\_length}. The model handles the input in one token sequence where sequence refers to the input to the ALBERT model which can be single sentence or multiple sentences. A special token [CLS] is appended at the beginning of every sequence. The final hidden state corresponding to this token is used in the small network to classify the emotion. Multiple sentences in the same token are separated by token [SEP]. A learned embedding token is added to each token to indicate the context which is passed through subsequent segments and positional embeddings. We use the pooled\_output of the pre-trained ALBERT model which has a size of { \it batch\_size, 768 } as the hidden size is 768. The output of the ALBERT model is then passed through a series of 3 Feedforward layers each with 100 nodes and reLu activation function which was concurred upon through the process of experimentation. The last layer is a dense layer which contains 6 nodes, each node to predict that specific emotion with a {\it softmax} activation function which smooths out our output accordingly on the IEMOCAP dataset.

\subsection{Computational Resources: \textit{Compute Canada}}
This research was enabled in part by support provided by \textit{Compute Canada}\footnote{www.computecanada.ca}
For running our codes, we use the \textit{Compute Canada} cluster, running on 14 CPU cores with 30,000M CPU memory and allocated time for the execution of whole program as 150 minutes. The program terminates after 150 minutes or earlier with the output stored in a file. The GPU specifications used are as follows: Model: P100-PC21E-12GB, with two GPUs per CPU socket.

\section*{Results and Discussion}
\subsection{Comparison with Baseline and State-of-the-art model}
For comprehensive evaluation of our model, a comparison is made with the baseline models and state-of-the-art models. 
We provide a thorough comparison of results mentioned in Table \ref{comparison} with the baseline-models as well as the state of the art models. We use both the accuracy and Weighted F1-Score for comparison between various models.
Following is a brief description of each model used for comparison:
\begin{itemize}
    \item CNN: The model architecture presented by \cite{Kim2014} is variant of the traditional Convolutional neural networks (CNN) architecture. The feature extractor focuses on capturing the important feature, the one with the highest value obtained from the feature map. The multiple filter feature extractor does not use contextual information and works with variable sentence lengths.

\item Memnet: A neural network in the form of memory network which is trained end-to-end and hence requires less supervision is proposed. The sentences in the model presented by \cite{Sukhbaatar2015} are represented in two different representations. The first approach is bag of words which embeds each word of the sentence and sums the vectors which in turn fails to capture the importance of key words. The second approach focuses more on the position of the words wherein element wise multiplication is performed in order to obtain the positional encoding. The main essence of the model lies in the importance of the temporal context of the sentence. To enable this in the model, the memory vector is modified to encode the temporal information. The output matrix is also embedded the same way. Both the matrices are learned during training, which helps in getting the best possible answer that is the emotion behind the sentences.

\item CMN: The proposed conversational memory networks (CMN) model by \cite{hazarika-etal-2018-conversational} involves multimodal feature extraction for all utterances involving memory networks. The process of feature extraction involves a fusion of features extracted at 3 levels - audio, visual, and textual which are joined to form the utterance representation. The memory network is modeled using GRU cells which considers the context information from the history and its preceding utterances. Finally, the dynamics of the model are presented by feeding the current utterance to two distinct memory networks.

\item DialogueRNN: \cite{Majumder2018} proposed a system which employs three gated recurrent units focusing on the three major aspects of emotion detection: the speaker, context from preceding emotions and emotions behind the preceding utterances. The proposed model follows feature extraction technique similar to CNN followed by the three GRU layers.

\item BiDialogueRNN: \cite{Majumder2018} proposed another variant of DialogueRNN which has both a forward pass and a backward pass using RNN. In this variant, the emotion is passed taking into consideration both the previous utterances and the upcoming utterances.

\end{itemize}

As evident, AdCOFE surpasses the accuracy and F1-score of all the baseline models (CNN, Memnet, and CMN) and the state-of-the-art Dialogue RNN model. In terms of F1 score, AdCOFE surpasses the accuracy of baseline models approximately by 16.5\%, 9.6\% and 8.6\%  respectively. When compared to BiDialogueRNN model AdCOFE exceeds the accuracy and F1 score approximately by 1.1\% and 2\% respectively.

\begin{center}
\begin{table*}
\footnotesize  
  \begin{tabular}{p{1.75cm}| p{0.48cm}| p{0.48cm}| p{0.48cm}| p{0.48cm}| p{0.48cm}| p{0.48cm}| p{0.48cm}| p{0.48cm}| p{0.48cm}| p{0.48cm}| p{0.48cm}| p{0.48cm}| p{0.48cm}| p{0.48cm} }
    \hline
      {\textbf{Methods}} &
      \multicolumn{2}{c}{\textbf{Happy}}&
      \multicolumn{2}{c}{\textbf{Sad}}&
      \multicolumn{2}{c}{\textbf{Neutral}}&
      \multicolumn{2}{c}{\textbf{Angry}}&
      \multicolumn{2}{c}{\textbf{Excited}}&
      \multicolumn{2}{c}{\textbf{Frustrated}}&
      \multicolumn{2}{c}{\textbf{Average}}\\
    
     & \textbf{Acc} & \textbf{F1} & \textbf{Acc} & \textbf{F1} & \textbf{Acc} & \textbf{F1} & \textbf{Acc} & \textbf{F1} & \textbf{Acc} & \textbf{F1} & \textbf{Acc} & \textbf{F1} & \textbf{Acc} & \textbf{F1} \\
   \hline
    CNN & 27.77 & 29.86 & 57.14 & 53.83 & 34.33 & 40.14& 61.17& 52.44& 46.15& 50.09& 62.99& 55.75& 48.92& 48.18 \\
    
    Memnet & 25.72 & 33.53 & 55.53 & 61.77 & 58.12 & 52.84 & 59.32 & 55.39 & 51.50 & 58.30 & 62.70 & 59.00 & 55.72& 55.10\\
    
     CMN & 25.00 & 30.38 & 55.92 & 62.41 & 52.86 & 52.39 & 61.76 & 59.83 & 55.52 & 60.25 & 71.13 & 60.69 & 56.56& 56.13\\
    
    DialogueRNN & 28.47 & 36.61 & 65.31 & 72.40 & \textbf{62.50} & 57.21 & 67.65& 65.71& 70.90& 68.61& 61.68& 60.80& 61.80& 61.51 \\
    
    BiDialogueRNN & 25.69 & 33.18 & \textbf{75.10} & \textbf{78.80} & 58.59 & 59.21 & 64.71 & 65.28& \textbf{80.27}& \textbf{71.86}& 61.15& 58.91& 63.40& 62.75\\
    
    \textbf{AdCOFE} & \textbf{54.94} & \textbf{54.84} & 56.69 & 56.64 & 61.73 & \textbf{59.68} & \textbf{72.71} & \textbf{73.04} & 64.11& 65.00& \textbf{69.67}& \textbf{67.12} & \textbf{64.51} & \textbf{64.72}\\
    \hline
  \end{tabular}
\caption{Comparison with baseline models\label{comparison}}
\end{table*}
\end{center}

\subsection{Comparison with latest ERC models}
The research in the field of Emotion Recognition has been rapidly increasing and to provide a comprehensive idea of AdCOFE model's performance, a comparison of the results is made with the following latest advanced ERC models:
\begin{itemize}
    \item EmoGraph : The proposed model consists of a graphical neural network which uses co-occurrence statistics in between every emotion classes considering each emotion class as a node of the graph. This network then helps in extracting features from neighbours of each emotion node \cite{Xu2020}.
    \item AGHMN: \cite{Jiao2019a} proposed an Attention Gated Hierarchical Memory Network (AGHMN) consisting of a the Hierarchical Memory Network (HMN) with primary focus on enhancement of features at utterance level and of contextual information memory bank. Two layers are defined for modeling of word sequence and for capturing historical utterances with Attentions GRU to retain positional information of the utterance.
    \item RGAT: \cite{Ishiwatari2020} created a Relation-aware Graph Attention network which  would  reflect  the  speaker  de-pendency and sequential information by applying positional encodings in the relational graph structure. 
\end{itemize}

\begin{table}[ht]
\centering
\begin{tabular}{l|r}
Model & F1 score \\\hline
EmoGraph & 65.4\\
AGHMN  & 63.5\\
RGAT   & 65.22\\
\textbf{AdCOFE}   & 64.7\\
\end{tabular}
\caption{Comparison with latest ERC models\label{latest}}
\end{table}

AdCOFE is heavily based on extracting contextual features from the data in contrast to current complex models using graphical and attention based neural network and provide comparable results (Table \ref{latest}).

\subsection{Model Variants}
The following section deals with the different variants of the framework presented in this paper.The results of each variant along with their respective F1 score are displayed in the (Table \ref{variants}) below.

Initially the fine tuned data is passed to the ALBERT model. The performance obtained in terms of classification of emotions is low as the model doesnot consider contextual information. The model treats the data as just lines of conversations and fails to capture high level contextual information. This results in low accuracy. Further to enhance the accuracy the sentences in the dataset are arranged in specific batches wherein similar sentences of one speaker are clubbed together. This variation focuses on passing on the emotion throughout which in turn improves the accuracy. 
Since any conversation associates a natural meaning along with it, more focus on the meaning of each word helps us in better classification of emotions. The improvised version of the model makes use of knowledge graph which capture the natural language meaning behind each word and sentences. The results generated as a result of the ConceptNet API further enhances the contextual information of the utterances. The final variant combines all the above features along with sentiment lexicon which rates the sentences according to the valence as positive, negative, neutral based on a threshold. This added embedding extensively improves the accuracy. The final product of provides us with  a framework which overcomes all the disadvantages of the previous old models and majorly focuses on feature extraction and contextual enrichment.

\begin{table}[ht]
\centering
\begin{tabular}{l|r|c}
Model & Accuracy & F1 score \\\hline
ALBERT & 58.2  & 58.6\\
ALBERT + Batch Sentences & 58.4  & 58.7\\
ALBERT + ConceptNet & 62.3  & 62.7\\
ALBERT + ConceptNet +VADER & 64.1  & 64\\
\end{tabular}
\caption{Variants of Model\label{variants}}
\end{table}

\section*{Conclusion and Future Work}
AdCOFE focuses heavily on extracting contextual features from the data. The human brain is able to comprehend the emotion of the sentences not only based on the words and their relation in the sentence, but also because it has a vast contextual information from its own memory. We apply the same logic to execute our model by using knowledge graphs for advanced contextual understanding of the sentences and sentiment lexicons for adding context-based emotional features to the sentences. The implemented model surpasses the accuracies and F1-score for the baseline models and the state-of-the-art model. Furthermore, AdCOFE entirely runs using pretrained models and transfer learning, and results are achieved by running only 4 epochs, thus making it a simple and elegant model to implement. Overall, AdCOFE implements a unique idea of focusing on context and accomplishes good results when compared with other complex neural network models. For future work, we plan to focus more on enhancing classification model for passing the emotions of each utterance throughout the conversation for capturing long-range contextual information more effectively.

\section*{Acknowledgement}
Publication costs are funded by NSERC Discovery Grant (RGPIN-2017-05377), held by Dr. Vijay Mago.

\bibliography{new_references}

\end{document}